\pdfoutput=1

\documentclass[runningheads]{llncs}
\usepackage{graphicx}

\usepackage{tikz}
\usepackage{comment}
\usepackage{amsmath,amssymb} 
\usepackage{booktabs}
\usepackage{enumitem}
\usepackage{makecell}
\usepackage{array}
\usepackage{color}
\usepackage{siunitx}
\usepackage{multicol}
\usepackage{multirow}
\usepackage{forloop}
\usepackage{xspace}
\usepackage{wrapfig}
\usepackage{tabularx}
\usepackage{arydshln}
\usepackage{subcaption}
\usepackage{calc}
\usepackage{orcidlink}

\usepackage{cprotect}
\makeatletter
\DeclareRobustCommand\onedot{\futurelet\@let@token\@onedot}
\def\@onedot{\ifx\@let@token.\else.\null\fi\xspace}

\def\eg{\emph{e.g}\onedot}

\def\etc{\emph{etc}\onedot}

\def\etal{\emph{et al}\onedot}

\newcommand{\defcal}[1]{\expandafter\newcommand\csname 
	c#1\endcsname{{\mathcal{#1}}}}
\newcommand{\defbb}[1]{\expandafter\newcommand\csname 
	b#1\endcsname{{\mathbb{#1}}}}
\newcommand{\defbf}[1]{\expandafter\newcommand\csname 
	bf#1\endcsname{{\mathbf{#1}}}}
\newcounter{calBbCounter}
\forLoop{1}{26}{calBbCounter}{
	\edef\letter{\Alph{calBbCounter}}
	\expandafter\defcal\letter
	\expandafter\defbb\letter
	\expandafter\defbf\letter
}    
\usepackage[capitalize]{cleveref}
\crefname{section}{Sec.}{Secs.}
\Crefname{section}{Section}{Sections}
\Crefname{table}{Table}{Tables}
\crefname{table}{Tab.}{Tabs.}
\newcommand{\RomanNumeral}[1]{\romannumeral #1}
\newcommand{\RomanNumeralCaps}[1]{\uppercase\expandafter{\romannumeral#1}}

\newcommand{\FixedMM}[1][]{\cI^{#1}}
\newcommand{\DynamMM}[1][]{m^{#1}}

\newcommand{\AudioItem}[1]{a_{#1}}
\newcommand{\AudioSeq}{\cA_t^s}
\newcommand{\AudioFeat}[1]{s_{#1}}
\newcommand{\AudioFeats}{\cS_t^s}

\newcommand{\SpeakerItem}[1]{v_{#1}^s}
\newcommand{\SpeakerSeq}[1][t]{\cV_{#1}^s}

\newcommand{\SpeakerFeat}[1]{m_{#1}^s}
\newcommand{\SpeakerFeats}{\cM_t^s}

\newcommand{\ListenerItem}[1]{v_{#1}^l}
\newcommand{\ListenerSeq}[1][t]{\cV_{#1}^l}

\newcommand{\ListenerFeat}[1]{m_{#1}^l}

\newcommand{\ListenerFeatstPLUS}{\cM_{t+1}^l}
\newcommand{\ListenerFeatsT}{\cM_T^l}
\newcommand{\ListenerFeatHat}[1]{\hat{m}_{#1}^l}

\newcommand{\ListenerFeatsHatT}{\hat{\cM}_T^l}

\newcommand{\Generator}{\bfG_m}
\newcommand{\Render}{\bfG_v}

\newcommand{\Dataset}{\cD}
\newcommand{\TrainSet}{\Dataset_{train}}
\newcommand{\EvalSet}{\Dataset_{test}}
\newcommand{\OODSet}{\Dataset_{ood}}
\newcommand{\Attitude}{e}

\usepackage[accsupp]{axessibility}  
\definecolor{apositive}{RGB}{248, 203, 172}
\definecolor{anegative}{RGB}{180, 199, 231}
\definecolor{aneutral}{RGB}{197, 224, 181}

\usepackage{pifont}
\newcommand{\cmark}{\ding{51}}%
\newcommand{\xmark}{\ding{55}}%

\usepackage[accsupp]{axessibility}  

\begin{document}
\pagestyle{headings}
\mainmatter
\def\ECCVSubNumber{6056}  

\title{Responsive Listening Head Generation:\\ A Benchmark Dataset and Baseline}


\titlerunning{Responsive Listening Head Generation: A Benchmark Dataset and Baseline}
%
\author{Mohan Zhou$^{1*\dagger}$\orcidlink{0000-0003-3250-4978} \and
Yalong Bai$^{2\dagger}$\orcidlink{0000-0002-8416-9027} \and
Wei Zhang$^{2}$\orcidlink{0000-0002-1492-8286} \and
Ting Yao$^{2}$\orcidlink{0000-0001-7587-101X} \and
Tiejun Zhao$^{1\S}$\orcidlink{0000-0003-4659-4935} \and
Tao Mei$^{2}$\orcidlink{0000-0002-5990-7307}}

\authorrunning{M. Zhou et al.}
%
\institute{$^1$Harbin Institute of Technology\qquad $^2$JD Explore Academy, Beijing, China\\
\email{\{mhzhou99, ylbai\}@outlook.com, \{wzhang.cu, tingyao.ustc\}@gmail.com, tjzhao@hit.edu.cn, tmei@jd.com}
\footnotetext{$*$~This work was done at JD Explore Academy.}
\footnotetext{$\dagger$~Equal contribution.~~~$\S$ Corresponding author.}
}
\maketitle

\begin{abstract}
We present a new listening head generation benchmark, for synthesizing responsive feedbacks of a listener (\eg, nod, smile) during a face-to-face conversation. As the indispensable complement to talking heads generation, listening head generation has seldomly been studied in literature.
Automatically synthesizing listening behavior that actively responds to a talking head, is critical to applications such as digital human, virtual agents and social robots.
In this work, we propose a novel dataset ``ViCo'', highlighting the listening head generation during a face-to-face conversation. A total number of 92 identities (67 speakers and 76 listeners) are involved in ViCo, featuring 483 clips in a paired ``speaking-listening" pattern, where listeners show three listening styles based on their attitudes: positive, neutral, negative. Different from traditional speech-to-gesture or talking-head generation, listening head generation takes as input both the audio and visual signals from the speaker, and gives non-verbal feedbacks (\eg, head motions, facial expressions) in a real-time manner. Our dataset supports a wide range of applications such as human-to-human interaction, video-to-video translation, cross-modal understanding and generation. To encourage further research, we also release a listening head generation baseline, conditioning on different listening attitudes. Code \& ViCo dataset: \url{https://project.mhzhou.com/vico}.


\end{abstract}

\section{Introduction}
\label{sec:intro}
Communication~\cite{berger2005interpersonal,honeycutt2001mental,luhmann1992communication,parker2000improving,stacks2014integrated,tomasello2010origins} is one of the most common activities that everybody engages in their daily lives. During a face-to-face communication~\cite{kendon2011organization}, two persons shift their roles in turn between the speaker and listener, to effectively exchange information. The speaker verbally transmits information to the listener, while the listener provides real-time feedbacks to the speaker mostly through non-verbal behaviors such as \textit{affirmative nod}, \textit{smiling}, \textit{head shake}. 

\begin{figure*}[t]
 \centering
  \includegraphics[width=0.92\linewidth]{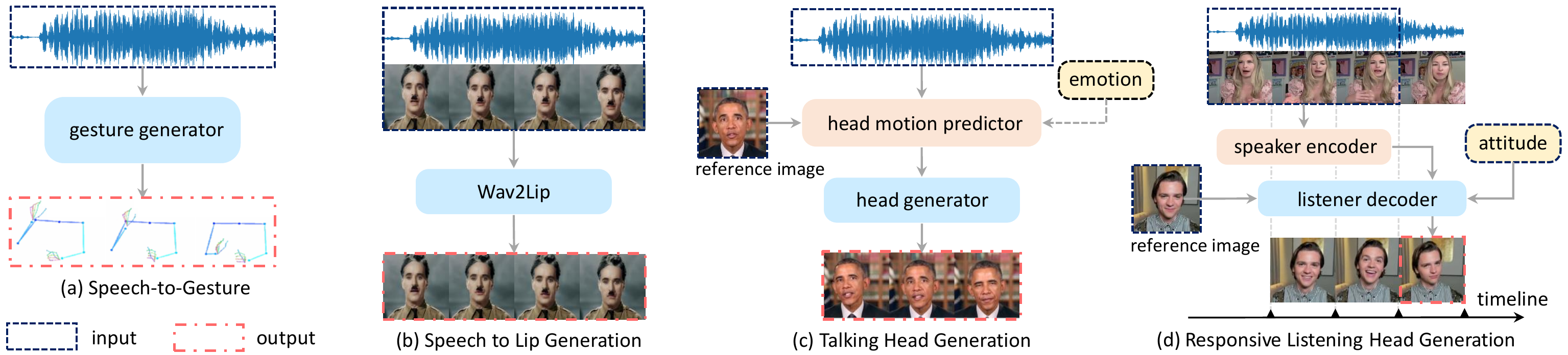}
 \caption{Illustrations of three related tasks and our proposed responsive listening head generation. (a) Speech-to-gesture translation: generates plausible gestures to go along with the given speech. (b) Speech to lip generation: produces lip-synchronization in talking-head video. (c) Talking head generation: synthesizes talking face video conditioned on the identity of the speaker, audio speech, and/or the speaker emotion. (d) Our proposed responsive listening head synthesizes videos in responding to the speaker video stream}
  \label{fig:taskdescription}
\end{figure*}

Although static images, repeated frames, or pre-scripted animations are often used to synthesize listeners in practice, they are often rigid and not realistic enough to respond to the speaker appropriately. According to studies in social psychology and anthropology, listening is a function-specific~\cite{hadar1985head} and conditioned behavior~\cite{barker1971listening}, where learnable patterns can be inferred from training data. First, common patterns of listeners are observed to express their viewpoints, symmetrical and cyclic motions were employed to signal ‘yes’, ‘no’ or equivalents; narrow linear movements occurred in phase with stressed syllables in the other's speech; wide, linear movements occurred during pauses in the other's speech. Even the duration of eye blinks of the listener is perceived as communicative signals in human face-to-face interaction~\cite{homke2018eye}. 
Second, these patterns in listener motions are mainly affected by two signals: the attitude of listener~\cite{heylen2007searching}, and signals from the speaker~\cite{cassel2000elements,gillies2008responsive,maatman2005natural}. Different attitudes of the listener results in diverse facial expressions, \eg., attitude of \textit{agree} is meant by a \textit{nod} and \textit{accept}, attitude of \textit{disbelieve} is represented by the combination of \textit{head tilt} and \textit{frown}. Meanwhile, listening behavior is heavily affected by speaker motion and audio signals. For example, the flow of movement of listener may be rhythmically coordinated with the speech and motions by the speaker~\cite{kendon1970movement}. These psychological and ethological studies motivate us to propose a data-driven method for modeling listening behaviors for face-to-face communication.

There have been extensive research efforts on speaker-centric synthesis. As shown in \cref{fig:taskdescription}, speech to gesture generation~\cite{ginosar2019learning} learns a mapping between the audio signal and speaker's pose. Speech to lip generation~\cite{prajwal2020lip} aims to refine the lip-synchronization of a given video input. Talking-head synthesis~\cite{chung2017you,wang2020mead,zhu2021deep} tries to generate a vivid talking video of a specific speaker with facial animations from a still image and a clip of audio. However, these works only focus on the speaking role, while ignore the indispensable counterpart of listener. Notably during a face-to-face conversation, listening behavior is even more important, as proper feedbacks to the speaker (\eg, \textit{nod}, \textit{smile}, \textit{eye contact}, \etc.) are vital for a successful communication~\cite{mcnaughton2008learning,robertson2005active,rogers1957active,rost2013active}. Through real-time feedbacks, listeners show how they are engaged (\eg, \textit{interested}, \textit{understand}, \textit{agree}, \etc.) to the speech, such that conversation gets more accessible for both participates. 

In this work, we propose a new task to highlight listener-centric generation. Specifically, listening-head generation aims to synthesize a video of listening head, conditioning on the corresponding talking-head video of the speaker and the identity information of the listener, as shown in \cref{fig:taskdescription}d. Proper reactions of the listener are expected to coordinate with the input talking video. This task is critical to a wide range of applications including virtual anchors, digital influencers, customer representatives, digital avatar in Metaverse, wherever involves interactive communication.

To address this, we construct a high-quality speaker-listener dataset, named ViCo, by capturing the high-definition video data from public conversations between two persons containing frontal faces on the same screen. The data strictly follows the principle that a video clip contains only uniquely identified listener and speaker, and requires that the listener has responsive non-verbal feedback to the speaker. After data cleaning, we further annotate the listener with three different attitudes: positive, neutral and negative. In total, our ViCo dataset contains 483 video clips of 76 listeners responding to 67 speakers. Compared to speaker-centric datasets such as MEAD~\cite{wang2020mead} and VoxCeleb2~\cite{chung2018voxceleb2}, ViCo highlights the listener role, making an indispensable couterpart to those speaker-centric ones. Compared to SEMAINE~\cite{mckeown2011semaine} (human interacts with a limited artificial agent) and MAHNOB Laughter~\cite{petridis2013mahnob} (people watching movies), ViCo features real persons in real conversations, such that natural reactions between genuine humans during a conversation make a key difference.

Together with the dataset, we propose a listening-head generation baseline method. We are aware that previous speaker-centric tasks are usually modeled in an idiosyncratic way (different speakers are modeled independently). However, listening behavior patterns are typically well coordinated with the speaker video. Thus we decouple the identity features from the listener and focus on learning the general motion patterns of responsive listening behaviors. We model listening head generation as a video-to-video translation task, by designing a sequence-to-sequence architecture to sequentially decode the listener's head motion and expression. Through quantitative evaluation and user study, we show our baseline is able to automatically capture the salient moments of speaker video and responds properly with clear motions and expressions.
\section{Related Works} \label{sec:related}
\textbf{Active listener}\quad Active listening is an effective communication skill that not only means focusing fully on the speaker but also actively showing the non-verbal signals of listening with attitude. Usually, the active listener would mirror some facial expressions used by the speaker or shows more eye contact with the speaker. Active listening have shown its positive effects in many areas, such as teaching~\cite{jalongo1995promoting}, medical consultations~\cite{fassaert2007active}, team management~\cite{mineyama2007supervisors}, \etc. In this paper, we aim to generate an active listener that could provide responsive feedback, the listener would understand the speaker’s verbal and non-verbal signals first and then give proper feedback to the speaker.

\textbf{Speaker-centered video synthesis}\quad Given time-varying signals and a reference still image of the speaker, the talking head synthesis task aims to generate a vivid clip for the speaker with the time-varying signals matched. Based on the different types of time-varying signals, we can group these tasks into two groups: 1) audio-driven talking head synthesis~\cite{chung2017you,prajwal2020lip,zhang2021facial}, 2) video-driven talking head synthesis~\cite{bansal2018recycle,wu2018reenactgan}. The goal of the former one is to generate a video of the speaker that matches the audio. And the latter one is to generate videos of speakers with expressions similar to those in the video. This differs from our task: the ``listener'' is forced to perceive the speaker's visual and audio signals and make an active response. Our task does not focus on only a single person or transfers face expression and slight head movements from another person. There are two roles in our task: listener and speaker, and the listener should actively respond to the speaker with non-verbal signals.

\begin{table}[b]
  \centering
  \small
  \caption{Comparison with other listener-related datasets}
  \label{tab:dataset_comparison}
  \begin{tabular}{@{}lcccc@{}}
    \hline\noalign{\smallskip}
    Dataset & Public & Environment & Style & Interact with Real \\
    \noalign{\smallskip}
    \hline
    \noalign{\smallskip}
    Gillies~\etal~\cite{gillies2008responsive} & \xmark & Lab & Simulated & \xmark \\
    SEMAINE~\cite{mckeown2011semaine} & \cmark & Lab & Simulated & \cmark \\
    Heylen~\etal~\cite{heylen2011generating} & \xmark & Lab & Simulated & \cmark \\
    MAHNOB Laughter~\cite{petridis2013mahnob} & \cmark & Lab & Realistic & \xmark \\
    ALICO~\cite{buschmeier2014alico} & \xmark & Lab & Realistic & \cmark \\
    \hdashline[.4pt/1pt]
    Ours & \cmark & Wild & Realistic & \cmark \\
    \hline
  \end{tabular}
\end{table}

\textbf{Listening behaviors modeling}\quad Many applications and research papers have focused on speaking, while the ``listener modeling'' is seldomly explored. Gillies \etal~\cite{gillies2008responsive} first propose the data-driven method that can generate an animated character that can respond to speaker's voice. This lacks the supervision of speaker visual signals, which is incomplete for responsive listener modeling. And this method can not be applied to realistic head synthesis. Heylen \etal~\cite{heylen2011generating} further studied the relationship between listener and speaker audio/visual signals from a cognitive technologies view. SEMAINE~\cite{mckeown2011semaine} records the conversation between a human and a limited artificial listener. MAHNOB Laughter database~\cite{petridis2013mahnob} focuses on studying laughter's behaviors when watching funny video clips. Apart from these related work, ALICO~\cite{buschmeier2014alico} corpus about active listener analysis is the most relevant dataset with our proposed task. However, it has not been made public and also not constructed from the real scene conversations. Moreover, the main objective of ALICO is for psychology analysis, the data mode of that dataset is vastly different from the audio-video corpora in computer vision area. In the past few years, the social AI intelligence~\cite{joo2019towards,oertel2021towards} has been introduced to model the nonverbal social signals in triadic or multi-party interactions. Joo~\etal~\cite{joo2019towards} concerned with the overall posture and head movement of a person, and Oertel~\etal~\cite{oertel2021towards} aims to mine listening motion rules for robotics controlling. Both related works only deal with the speaking status and ignore the speaker's content. What's more, they rarely care about two-person interactions nor pay attention to model the face in detail, which is also different from our task. A detailed comparison to exising listener-related datasets is shown in~\cref{tab:dataset_comparison}.

As far as we know, this is the first time to introduce the learning-based listening head generation task in computer vision area. In this work, we propose a formulation of responsive listening head generation and construct a public ViCo dataset for this task. Meanwhile, a baseline method is proposed for listening head synthesis by perceiving both speaker's audio/visual signals and preset attitude.
\section{Task Overview} \label{sec:overview}
We present Responsive Listening Head Generation, a new task that challenges vision systems to generate listening heads actively responding to the speaker's face or/and audio in real-time. In particular, we need to understand the head motion, facial expression, including eye blinks, mouth movements, \etc, of the input speaker video frame, and simultaneously understand the speaker's voice, then synchronously generate the active listening face video conditioned by the given attitude.

Given an input video sequence $\SpeakerSeq=\{\SpeakerItem{1},\cdots,\SpeakerItem{t}\}$ of a speaker head in time stamps ranging from $\{1,...,t\}$, and an corresponding audio signal sequence $\AudioSeq=\{\AudioItem{1},\cdots,\AudioItem{t}\}$ of the speaker, listening head generation aims to generate a listener's head $\ListenerItem{t+1}$ of the next time stamp:
\begin{equation}
    \ListenerItem{t+1} = \bfG(\SpeakerSeq, \AudioSeq, \ListenerItem{1}, \Attitude),
\end{equation}
where $\ListenerItem{1}$ is the reference head of the listener, $\Attitude$ denotes the attitude of the listener. The whole generated listener video $\ListenerSeq[t+1]$ can be denoted as the concatenation of $\{\ListenerItem{2},\cdots,\ListenerItem{t+1}\}$.

\setlength\intextsep{0pt}
\begin{wrapfigure}[10]{r}{0.5\textwidth}
    \centering
    \includegraphics[width=0.5\textwidth,page=1]{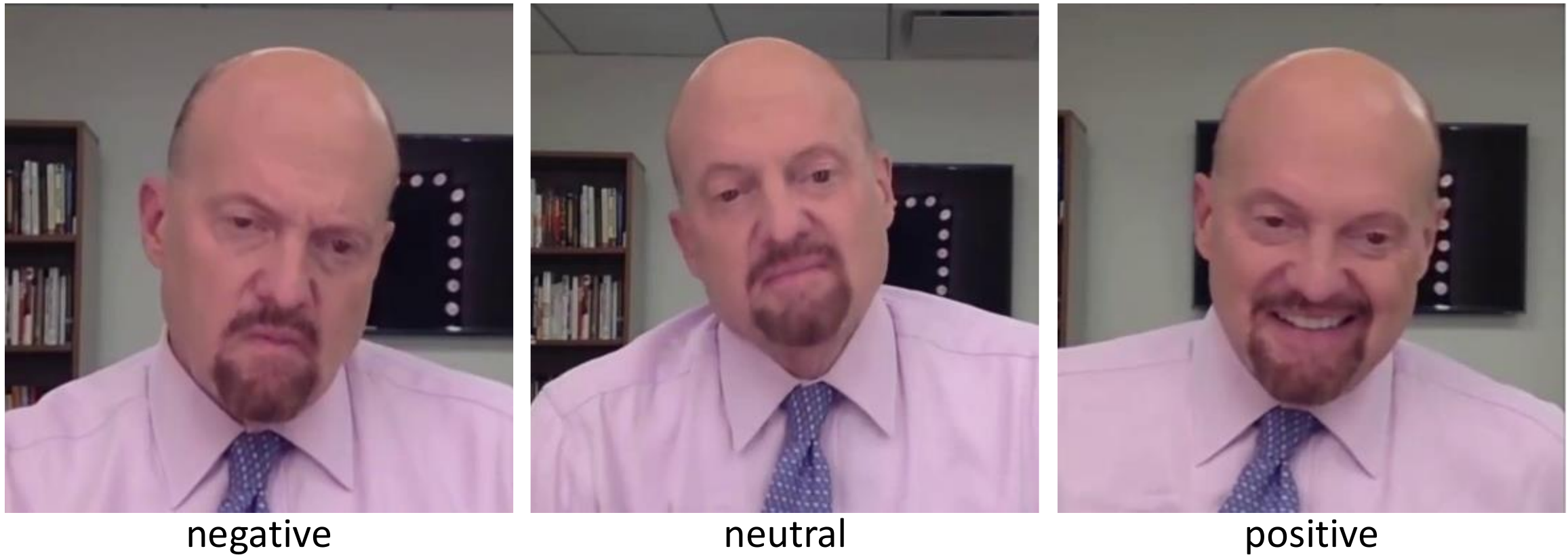}
    \caption{During a conversation, different attitudes of the listener could show different pose and expression patterns}
    \label{fig:attitude_example}
\end{wrapfigure}

\textbf{Listening attitude definition}\quad During conversation, after perceiving the signals from the speaker, the listener usually reacts with an active, responsive \emph{attitude}, including epistemic attitudes (\eg, agree, disagree) and affective attitudes (\eg, like, dislike). In this work, we group the attitudes into three categories: positive, negative and neutral. Positive attitude consists of \textit{agree}, \textit{like}, \textit{interested}. Conversely, negative attitude consists of \textit{disagree}, \textit{dislike}, \textit{disbelieve}, \textit{not interested}. In general, attitude potentially guides the listener's behavior and consequently affects the conversation. Also, different attitude results in different facial expressions and behaviors of the listener~\cite{heylen2007searching}, e.g. a smile appears as the most appropriate signal for \textit{like}, a combination of smile and raise eyebrows could be a possibility for \textit{interested}, \textit{disagree} can be meant by a head shake, \textit{dislike} is represented by a frown and tension of the lips, \etc. A listener example with different attitudes is illustrated in \cref{fig:attitude_example}.

\textbf{Feature extraction}\quad
In this work, we extract the energy feature, temporal domain feature, and frequency domain feature of the input audio; and model the facial expression and head poses using 3DMM \cite{blanz1999morphable} coefficients.

For the audio, we extract the Mel-frequency cepstral coefficients (MFCC) feature with the corresponding MFCC Delta and Delta-Delta feature. Besides, the energy, loudness and zero-crossing rate (ZCR) are also embedded into audio features $\AudioFeat{i}$ for each audio clip $\AudioItem{i}$. The audio feature extracted from $\AudioSeq$ can be denoted as $\AudioFeats=\{\AudioFeat{1},\cdots,\AudioFeat{t}\}$.

We leverage the state-of-the-art deep learning-based 3D face reconstruction model~\cite{deng2019accurate} for the videos to get the 3DMM~\cite{blanz1999morphable} coefficients. Specially, for each image, we can get the reconstruction coefficients $\{\alpha, \beta, \delta, p, \gamma\}$ which denote the identity, expression, texture~\cite{cao2013facewarehouse,bfm09}, pose and lighting~\cite{ramamoorthi2001efficient}, respectively. Further, we distinguished the 3D reconstruction coefficients into two parts: $\FixedMM=(\alpha, \delta, \gamma)$ to represent relatively fixed, identity-dependent features, and $\DynamMM=(\beta, p)$ to represent relatively dynamic, identity-independent features. The identity-independent feature extracted from speaker videos can be denoted as $\SpeakerFeats=\{\SpeakerFeat{1},\cdots,\SpeakerFeat{t}\}$, where $\SpeakerFeat{i}\in\mathbb{R}^{1\times C_v}$ is the expression and pose feature of 3D reconstruction coefficients for the $i$-th frame $\SpeakerItem{i}$, where $C_v = |\beta| + |p|$. 

\textbf{Task definition}\quad To ignore identity-dependent features and learn general listener patterns that can be adapted to multiple listener identities, we use only the head motion and facial expression feature $m$ for responsive listening head generation model training, and then adapt the identity-dependent features $\FixedMM$ of different listener identities for visualization and evaluation. Thus our listening head synthesis task can be formulated as:
\begin{equation}\label{eq:gm_orignal}
\begin{split}
    \ListenerFeat{t+1} &= \Generator(\SpeakerFeats, \AudioFeats, \ListenerFeat{1}, \Attitude),\\
    \ListenerItem{t+1} &= \Render(\ListenerFeat{t+1}, \FixedMM[l], \ListenerItem{1}),
\end{split}
\end{equation}
where $\ListenerFeat{t+1}$ is the dynamic feature predicted for listener's head, and $\FixedMM[l]$ denotes the identity-dependent features of the given listener. In a real implementation, we use $T$ frame of speaker's audio and video for responsive listening head generation model training.

The 3D face rendering technology $\Render$ has been well studied in many recent related works~\cite{kim2018deep,zhang2021facial}. Moreover, the face rendering models are usually identity-specific, so one may need to train the rendering model separately for each identity for better performance. To highlight the properties of the interactive digital human synthesis task, and decouple the critical factor in this task, our proposed responsive listening head synthesis model primarily focuses on the motion-related and identity-independent 3D facial coefficients prediction task $\mathbf{G}_m$, and use the pretrained rendering  model~\cite{ren2021pirenderer} for simplified visualization.
\section{Dataset Construction}
\label{sec:dataset}

\begin{figure}[t]     
    \centering
    \includegraphics[width=\linewidth,page=1]{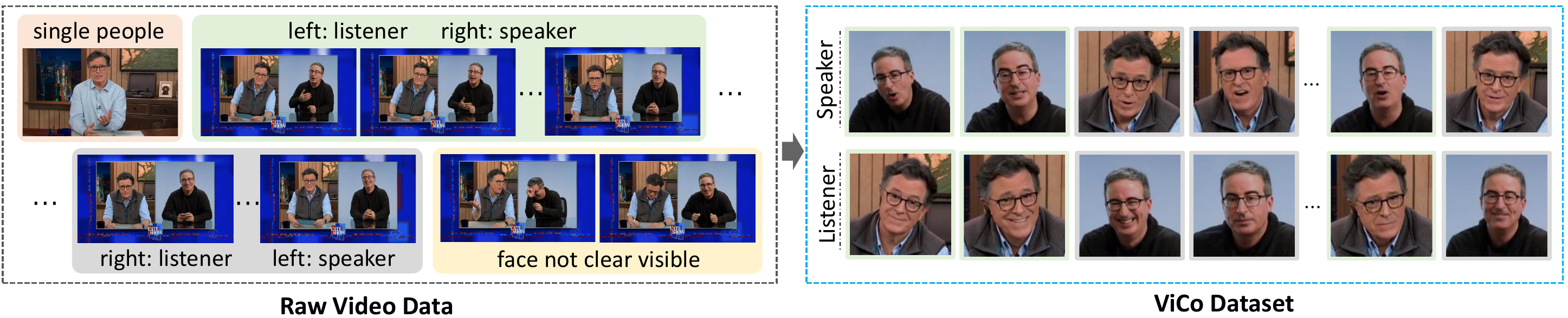}
    \caption{In ViCo, valid clips are selected in accordance with the standards that 1) both the speaker and listener behaviors are clearly visible, and 2) listeners are responsively engaged to the conversation. The facial regions of listener-speaker pairs are further cropped for constructing our ViCo dataset (right)}
    \label{fig:dataset_construct}
\end{figure}

We construct a dataset for responsive listening-head generation by capturing conversational video clips from YouTube containing two people's frontal faces. A \emph{valid} video clip is required to meet the following conditions: 
\begin{itemize}[itemsep=2pt,topsep=2pt,parsep=2pt]
    \item The screen contains only two people, and one of them is speaking while the other is listening carefully. %
    \item The frontal faces of both people are clearly visible. The facial expression is natural and stable.
    \item The listener actively responds to the speaker in a dynamic and real-time manner.
\end{itemize}

The annotators were asked to accurately record the start and end time of each \emph{valid} clip, label the position of the speaker (left or right of the screen) and identify the attitude of the listener in the video. Cross-validation was applied among at least three annotators for each candidate clip for quality control. For each valid clip, we use the MTCNN~\cite{zhang2016joint} to detect the face regions in each frame, and then crop and resize the detected face regions to 384$\times$384 resolution image sequence for model training and evaluation, as shown in~\cref{fig:dataset_construct}.

\setlength{\tabcolsep}{4pt}
\setlength\intextsep{8mm}
\begin{table}[t]
  \centering
  \small
  \cprotect\caption{Statistics of ViCo dataset. \verb|#|ID indicates the number of identities. The same person/identity can play different roles with multiple attitudes}
   \label{tab:dataset_statistics}
  \begin{tabular}{@{}lcccccc@{}}
    \hline\noalign{\smallskip}
    Attitude & \verb|#|Videos & \verb|#|Speaker & \verb|#|Listener & \verb|#|ID & \verb|#|Clips & Duration \\
    \noalign{\smallskip}
\hline
\noalign{\smallskip}
    Positive & 42 & 53 & 62 & 81 & 226 & \SI{49}{\minute}~\SI{18}{\second} \\
    Neutral  & 35 & 38 & 48 & 63 & 134 & \SI{27}{\minute}~~\SI{7}{\second} \\
    Negative & 11 & 11 &  9 & 18 &  123 & \SI{18}{\minute}~\SI{57}{\second} \\
    \hline
    Total    & 50 & 67 & 76 & 92 & 483 & \SI{95}{\minute}~\SI{22}{\second} \\
    \hline
  \end{tabular}
\end{table}
\setlength\intextsep   {8mm}

\Cref{tab:dataset_statistics} shows the statistical information of our annotated responsive listening head generation dataset ViCo. The proposed dataset contains rich samples of 483 video clips. We normalize all videos to 30 FPS, forming more than 0.1 million frames in total.  Moreover, our dataset has following properties:

\textbf{High quality}\quad All raw videos are of high resolutions (1920$\times$1080), so that the subtle differences between different attitudes and changing moods are well preserved. And audios are in \SI{44.1}{\kHz}/\SI{16}{bit} such that the speech-related features can be well preserved, too.

\textbf{High diversity}\quad Our dataset contains various scenarios, including news interviews, entertainment interviews, TED discussions, variety shows, \etc. These diverse scenarios provide rich semantic information and various listener patterns in different situations. The video clips length ranges from \SI{1} to \SI{71} seconds.

\textbf{Realtime and interactivity}\quad 
Different from the existing talking head video datasets~\cite{afouras2018lrs3,chung2018voxceleb2,li2021write,wang2020mead,zhang20213d} which aim to generate head or face in synchronization with the audio signals, our dataset focus on the face-to-face \emph{response}. These responses are generated by jointly understanding the speaker's audio, facial, and head motion signals, then adapting to different listener heads. It matters about mutual interaction rather than a monologue.
\begin{figure}[t]
    \centering
    \includegraphics[width=0.97\linewidth,page=1]{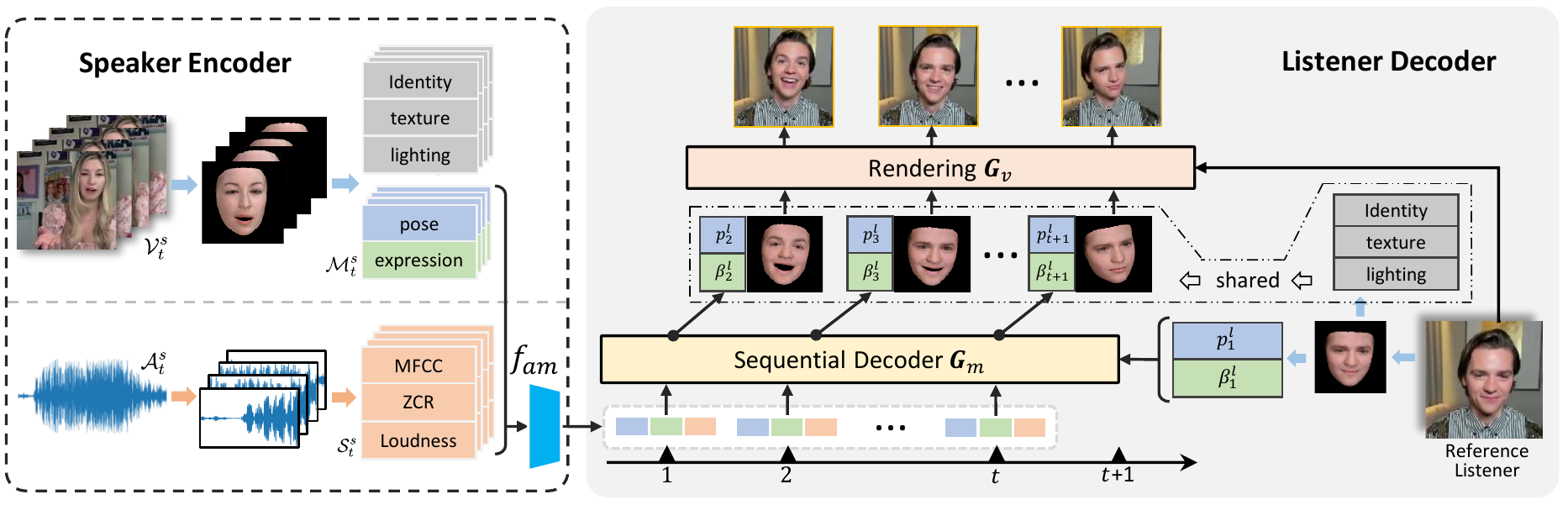}
    \caption{The overall pipeline of our responsive listening head generation baseline. The speaker encoder aims to encode the head motion, facial expression and audio features. Starting from the fused feature from reference listener image, the listener decoder receives signals from speaker encoder in temporal order, and predicts the head motion and facial expression features. These features are adapted to reconstruct the 3DMM coefficients with the reference listener's identity-dependent features, and then fed to a neural renderer to generate realistic listening video}
    \label{fig:model_arch}
\end{figure}
\section{Responsive Listening Head Generation} \label{sec:generation}
Based on ViCo dataset, we propose a responsive listening head generation baseline. The overview of our approach is illustrated in~\cref{fig:model_arch}. 

\subsection{Model architecture} \label{sec:model_arch}
According to the psychological knowledge, an active listener tends to respond based on speaker's audio~\cite{kendon1970movement} and visual signals~\cite{cassel2000elements,gillies2008responsive,maatman2005natural} comprehensively. And at a given moment, the listener receives information from the speaker of that moment as well as information from history and adopts a certain attitude to present actions in response to the speaker. Thus, the goal of our model is to estimate the conditioned probability $P(\ListenerFeatstPLUS \vert \SpeakerFeats, \AudioFeats, \ListenerFeat{1}, \Attitude)$, where the $\SpeakerFeats$ and $\AudioFeats$ are time-varying signals that the listener should respond to, and the reference listener feature $\ListenerFeat{1}$ and attitude $\Attitude$ constrain the pattern of the entire generated sequence.

Inspired by the sequence-to-sequence model~\cite{sutskever2014sequence}, a multi-layer sequential decoder module $\mathbf{G}_m$ is applied for modeling the time-sequential information of conversation. Unlike talking-head generation~\cite{chung2017you,wang2020mead,zhang2021facial,zhu2021deep}, which accepts an entire input of audio and then processes it using a bidirectional LSTM or attention layer; in our scenario, the model $\Generator$ receives the streaming input of the speaker where future information is not available.

For the speaker feature encoder, at each time step $t$, we first extract the audio feature $\AudioFeat{t}$ and the speaker's head and facial expression representation $\SpeakerFeat{t}$, then apply non-linear feature transformations following a multi-modal feature fusion function $f_{am}$ to get the encoded feature of speaker. The representation of reference listener $\ListenerFeat{1}$ and attitude $e$ can be embeded as the initial state $h_1$ for the sequential motion decoder. At each time step $t$, taking the speaker's fused feature $f_{am}(\AudioFeat{t}, \SpeakerFeat{t})$ as input, $\mathbf{G}_m$ in Eq.~\ref{eq:gm_orignal} is functioned as updating current state $h_{t+1}$ and generating the listener motion $m_{t+1}^l$, which contains two feature vectors, \textit{i.e.} $\beta_{t+1}^l$ for the expression and $p_{t+1}^l$ for the head rotation and translation. Our responsive listening head generator supports an arbitrary length of speaker input. The procedure can be formulated as:
\begin{equation}
    \beta_{t+1}^l, p_{t+1}^l = \mathbf{G}_m(h_t, f_{am}(s_t, m_t^s)).
\end{equation}

For optimization, with the ground truth listener patterns denoted as  $\ListenerFeatsHatT=[\ListenerFeatHat{2}, \ListenerFeatHat{3}, \cdots, \ListenerFeatHat{T}]$, we drop the last prediction $\ListenerFeat{T+1}$ due to the lack of supervision signals and use $L_2$ distance to optimize the training procedure:
\begin{equation}
    \cL_{gen} = \sum_{t=2}^{T} \Vert \beta_t^l - \hat{\beta}_t^l \Vert_2 + \Vert p_t^l - \hat{p}_t^l \Vert_2.
\end{equation}

Moreover, a motion constraint loss $\cL_{mot}$ is applied to guarantee the inter-frame continuity across $\ListenerFeatsHatT$ is similar to the predicted $\ListenerFeatsT$:
\begin{equation}
    \cL_{mot} = \sum_{t=2}^{T} w_1\Vert \mu(\beta_t^l) - \mu(\hat{\beta}_t^l) \Vert_2 + w_2\Vert \mu(p_t^l) - \mu(\hat{p}_t^l) \Vert_2,
\end{equation}
where $\mu(\cdot)$ measures the inter-frame changes of current frame and its adjacent previous frame, i.e., $\mu(\beta_t^l)=\beta_t^l-\beta_{t-1}^l$, $w_1$ and $w_2$ is a weight to balance the motion constraint loss and generation loss. The final loss function of our proposed listening head motion generation baseline can be formulated as:
\begin{equation}
    \cL_{total} = \cL_{gen} + \cL_{mot}.
\end{equation}

By optimizing $\cL_{total}$, our model can generate attitude conditioned responsive listening head for a given speaker video and audio.

\subsection{Implementation details} \label{sec:details}
To verify that our model can learn a generic listening pattern rather than conditioning on any particular individual, we divide the ViCo dataset ($\Dataset$) into three parts: \RomanNumeral{1}) training set $\TrainSet$ for learning listener patterns, \RomanNumeral{2}) test set $\EvalSet$ for validating our model on in-domain data, and \RomanNumeral{3}) out-of-domain (OOD) test set $\OODSet$ for evaluating the generalization and transferability. In this case, all identities in $\EvalSet$ have appeared in $\TrainSet$, while identities in the $\OODSet$ do not overlap with those in $\TrainSet$.

We extract 45-dimensional acoustic features for audios, including 14-dim MFCC, 28-dim MFCC-Delta, energy, ZCR and loudness. There are multiple choices to implement $\mathbf{G}_m$, such as standard sequential model like LSTM~\cite{hochreiter1997long}, GRU~\cite{chung2014empirical}, or a Transformer~\cite{vaswani2017attention} decoder with sliding window~\cite{beltagy2020longformer}. Here we adopt LSTM for our baseline, since it has been widely used in many similar applications such as motion generation~\cite{richard2021meshtalk}, and achieve stable state-of-the-art performance when training on small corpus~\cite{melis2019mogrifier}. Our listening head generation model is trained with AdamW~\cite{loshchilov2017decoupled} optimizer with a learning rate of \SI{1e-3}{} (decayed exponentially by 0.8 every 30 epochs), $\beta_1=0.9$ and $\beta_2=0.999$, for 300 epochs. For all experiments, we set hyper-parameter $w_1$ to 0.1 and $w_2$ to \SI{1e-4}{}.

\subsection{Experimental results} \label{sec:generation_results}

\textbf{Quantitative results}\quad Since we use a detached renderer rather than an end-to-end pipeline, we can divide the assessment into two sides: the performance of listener generator $\Generator$ and the visual effects of renderer $\Render$. For the former one, we use use $L_1$ distance between the generated features and the ground-truth features (FD) to ensure the predicted fine-grained head and expression coefficients similar to the ground-truth. And for the latter one, we select the Fr\'echet Inception Distance (FID)~\cite{heusel2017gans}, Structural SIMilary (SSIM)~\cite{wang2004image}, Peak Signal-to-Noise Ratio (PSNR) and Cumulative Probability of Blur Detection (CPBD)~\cite{bohr2013no} to evaluate the visual effects of renderer. The high-level metric on $\Generator$ can help us analyze the model, and the low-level metrics on $\Render$ provide a baseline for the successors.

\setlength{\tabcolsep}{4pt}
\begin{table}[t]
  \centering
  \small
  \caption{The Feature Distance ($\times 100$) of different listening head generation methods. Each cell in the table represent the feature distance of {\tt angle / expression / translation} coefficients respectively. Lower is better}
  \label{tab:fd_evaluate}
  \begin{tabular}{@{}lccccccccc@{}}
    \hline\noalign{\smallskip}
    \multirow{2}{*}{Attitude} & \multirow{2}{*}{Motion} & \multicolumn{2}{c}{Random} & \multicolumn{2}{c}{Simulation} & \multicolumn{2}{c}{Simulation$^*$} & \multicolumn{2}{c}{Ours} \\ \cline{3-10}
    \multicolumn{2}{c}{} & $\EvalSet$ & $\OODSet$ & $\EvalSet$ & $\OODSet$ & $\EvalSet$ & $\OODSet$ & $\EvalSet$ & $\OODSet$ \\ \noalign{\smallskip}
\hline
\noalign{\smallskip}
\multirow{3}{*}{Positive}  & angle & \makecell[r]{17.92} & \makecell[r]{17.99} & \makecell[r]{9.86} & \makecell[r]{10.48} & \makecell[r]{9.79} & \makecell[r]{11.57} & \makecell[r]{\textbf{6.79}} & \makecell[r]{\textbf{9.72}}\\
                          & exp & \makecell[r]{44.71} & \makecell[r]{44.86} & \makecell[r]{27.08} & \makecell[r]{27.81} & \makecell[r]{30.00} & \makecell[r]{30.27} & \makecell[r]{\textbf{15.37}} & \makecell[r]{\textbf{24.89}}\\
                          & trans & \makecell[r]{19.74} & \makecell[r]{20.14} & \makecell[r]{16.25} & \makecell[r]{12.06} & \makecell[r]{9.07} & \makecell[r]{13.82} & \makecell[r]{\textbf{6.48}} & \makecell[r]{\textbf{9.51}}\\
\hdashline[.4pt/1pt]
\multirow{3}{*}{Neutral}  & angle & \makecell[r]{17.85} & \makecell[r]{17.78} & \makecell[r]{10.94} & \makecell[r]{9.47} & \makecell[r]{14.18} & \makecell[r]{8.94} & \makecell[r]{\textbf{8.79}} & \makecell[r]{\textbf{6.33}}\\
                         & exp & \makecell[r]{44.26} & \makecell[r]{44.29} & \makecell[r]{27.37} & \makecell[r]{29.50} & \makecell[r]{26.44} & \makecell[r]{27.56} & \makecell[r]{\textbf{13.61}} & \makecell[r]{\textbf{23.51}}\\
                         & trans & \makecell[r]{19.98} & \makecell[r]{20.17} & \makecell[r]{8.47} & \makecell[r]{12.27} & \makecell[r]{11.53} & \makecell[r]{9.40} & \makecell[r]{\textbf{6.68}} & \makecell[r]{\textbf{8.95}}\\
\hdashline[.4pt/1pt]
\multirow{3}{*}{Negative}  & angle & \makecell[r]{19.53} & \makecell[r]{18.70} & \makecell[r]{17.86} & \makecell[r]{9.66} & \makecell[r]{13.24} & \makecell[r]{18.75} & \makecell[r]{\textbf{12.45}} & \makecell[r]{\textbf{8.54}}\\
                          & exp & \makecell[r]{45.68} & \makecell[r]{44.62} & \makecell[r]{29.57} & \makecell[r]{28.78} & \makecell[r]{31.62} & \makecell[r]{27.04} & \makecell[r]{\textbf{16.98}} & \makecell[r]{\textbf{18.99}}\\
                          & trans & \makecell[r]{19.69} & \makecell[r]{20.92} & \makecell[r]{8.06} & \makecell[r]{10.69} & \makecell[r]{24.42} & \makecell[r]{11.09} & \makecell[r]{\textbf{6.35}} & \makecell[r]{\textbf{5.81}}\\
\hline
\multirow{3}{*}{Average}  & angle & \makecell[r]{18.04} & \makecell[r]{18.11} & \makecell[r]{10.81} & \makecell[r]{9.91} & \makecell[r]{11.24} & \makecell[r]{12.58} & \makecell[r]{\textbf{7.79}} & \makecell[r]{\textbf{8.23}}\\
                         & exp & \makecell[r]{44.67} & \makecell[r]{44.60} & \makecell[r]{27.37} & \makecell[r]{28.66} & \makecell[r]{29.20} & \makecell[r]{28.46} & \makecell[r]{\textbf{15.04}} & \makecell[r]{\textbf{22.83}}\\
                         & trans & \makecell[r]{19.80} & \makecell[r]{20.36} & \makecell[r]{13.52} & \makecell[r]{11.76} & \makecell[r]{11.00} & \makecell[r]{11.55} & \makecell[r]{\textbf{6.52}} & \makecell[r]{\textbf{8.32}}\\
\hline
  \end{tabular}
\end{table}

\setlength{\tabcolsep}{4pt}
\begin{table}[t]
  \centering
  \small
  \caption{Quantitative Results of Renderer $\Render$ on $\EvalSet$ and $\OODSet$}
   \label{tab:render_performance}
  \begin{tabular}{@{}lcccccccc@{}}
    \hline\noalign{\smallskip}
    \multicolumn{1}{l}{\multirow{2}{*}{Attitude}} & \multicolumn{2}{c}{FID$\downarrow$} & \multicolumn{2}{c}{SSIM$\uparrow$} & \multicolumn{2}{c}{PSNR$\uparrow$} & \multicolumn{2}{c}{CPBD$\uparrow$} \\ \cline{2-9}
    \multicolumn{1}{c}{} & $\EvalSet$ & $\OODSet$ & $\EvalSet$ & $\OODSet$ & $\EvalSet$ & $\OODSet$ & $\EvalSet$ & $\OODSet$ \\ \noalign{\smallskip}
\hline
\noalign{\smallskip}
    Positive & 29.736 & 27.865 & 0.565 & 0.496 & 17.075 & 15.421 & 0.122 & 0.121 \\
    Neutral  & 36.551 & 27.366 & 0.686 & 0.544 & 21.220 & 17.703 & 0.106 & 0.113 \\
    Negative & 46.277 & 28.406 & 0.610 & 0.528 & 16.870 & 16.709 & 0.219 & 0.211 \\
    \hline
    Average  & 30.529 & 24.962 & 0.601 & 0.521 & 18.149 & 16.558 & 0.126 & 0.142 \\
    \hline
  \end{tabular}
\end{table}

\begin{figure}[t]
    \centering
    \includegraphics[width=0.98\linewidth,page=1]{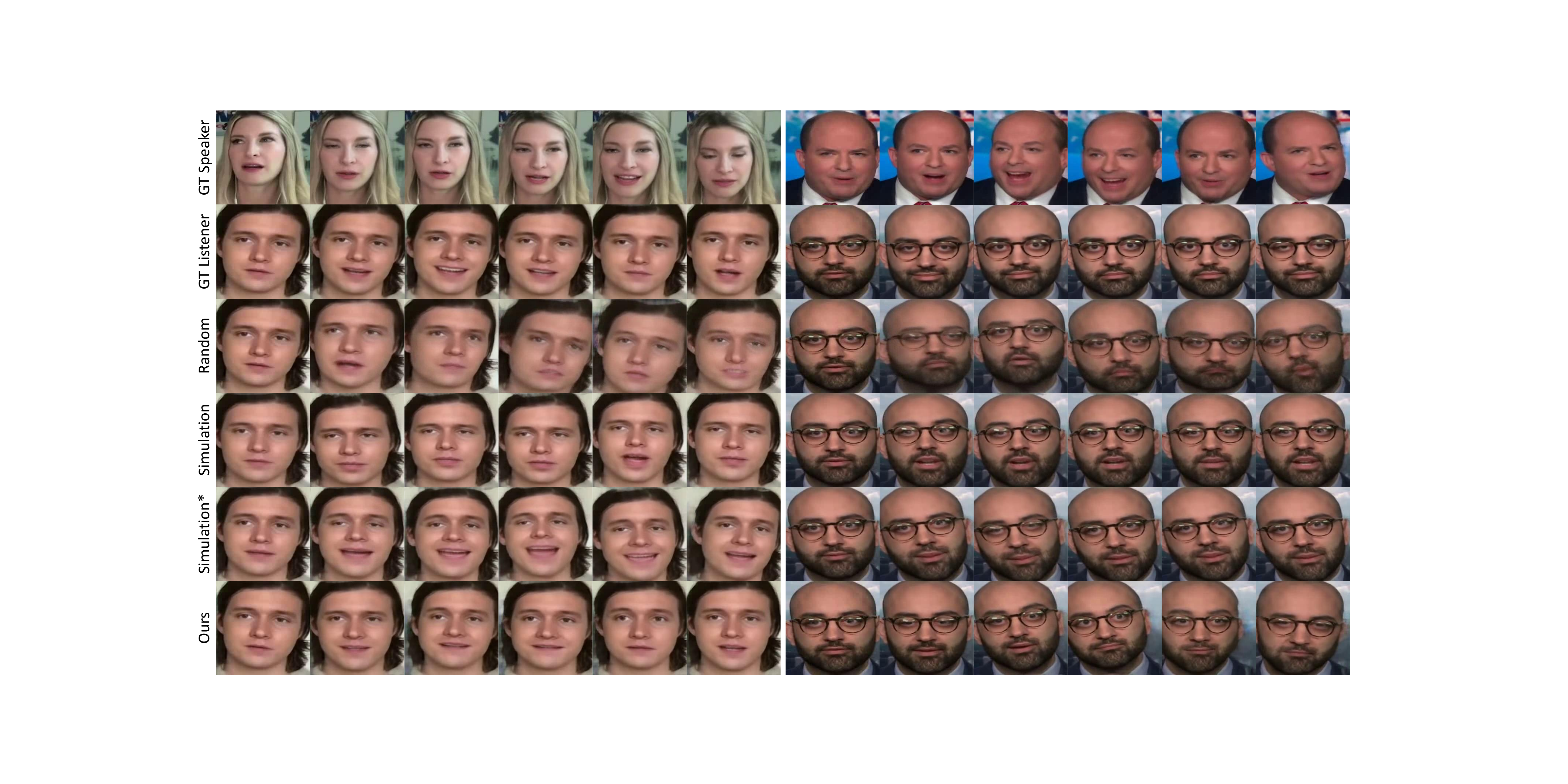}
    \caption{Qualitative comparison of listening head generation methods conditioned by the same reference frame (the first column of each group) and the same attitude (left: positive listener in $\EvalSet$, right: neutral listener in $\OODSet$). Our method can generate various, vivid and responsive listening motions for the given speaker video stream}
    \label{fig:qualitative_results}
\end{figure}

\begin{figure}[t]
    \centering
    \includegraphics[width=\linewidth]{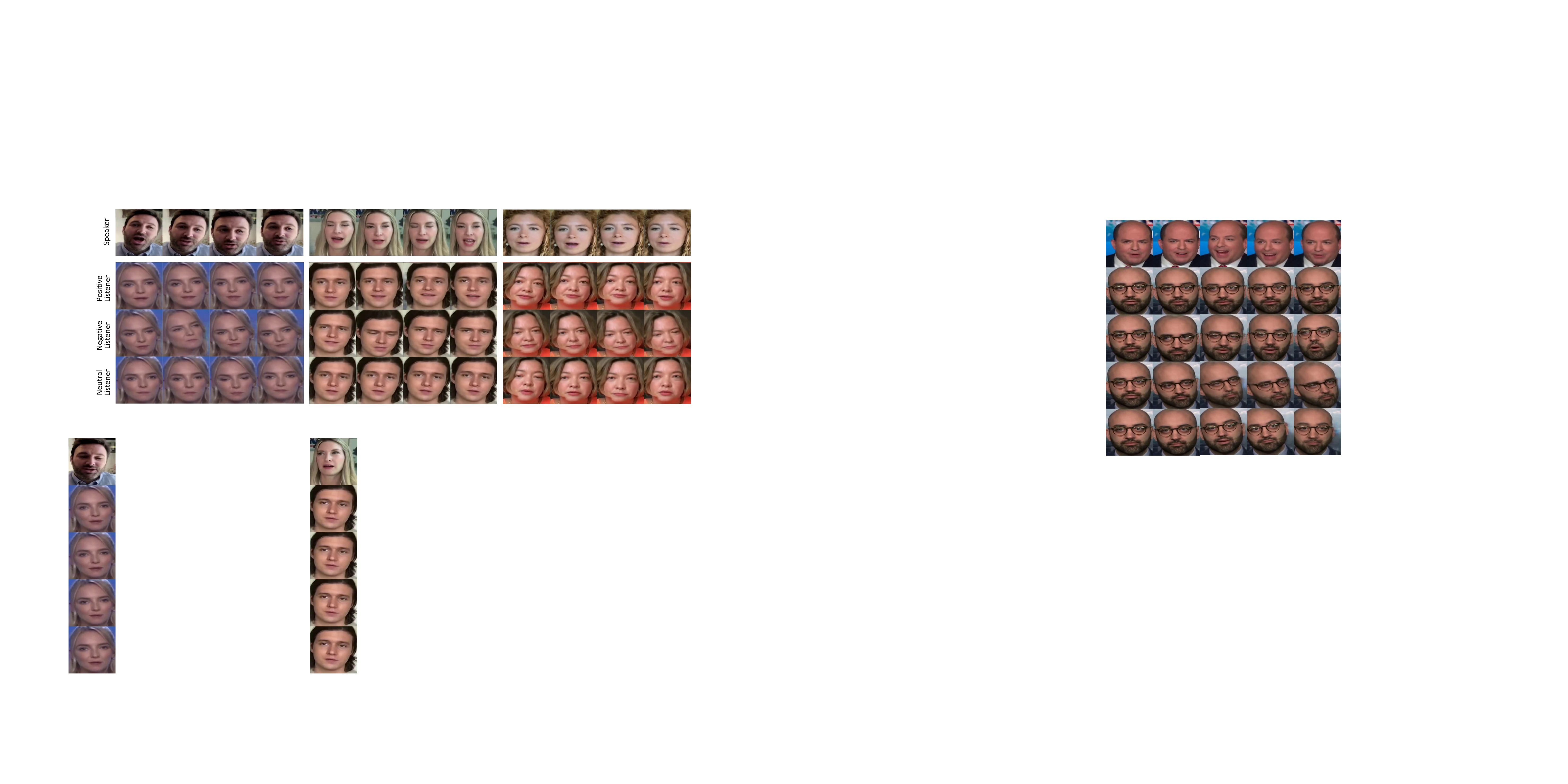}
    \caption{Comparative results of our listening heads generation method conditioned by the same reference image and speaker video but different attitudes}
    \label{fig:three_attitude}
\end{figure}

In~\Cref{tab:fd_evaluate}, we report the FD of the 3D facial coefficients across different listening head generation methods, including: 1) ``Random'': generate frames from reference image but injecting small perturbations in a normal distribution to mimic random head motion. 2) ``Simulation'': simulating natural listening behavior by repeating the motion patterns sampled from $\TrainSet$. 3) ``Simulation$^*$'': repeating the natural listening motion patterns sampled from $\TrainSet$ with the corresponding attitude. 4) ``Ours'': our proposed responsive listening head generation method.  The random permutations is the worst and not able to present a listener. Our method can reach the best performance which demonstrates the superiority of our algorithm over traditional non-parametric listeners. Also, we provide the evaluation results of zero-shot 3D face rendering~\cite{ren2021pirenderer} in \Cref{tab:render_performance}, as a basic criterion for evaluation of future realistic face rendering research work.

\textbf{Qualitative results}\quad Further, we visualize the results of those generations to analyze the differences between different configurations intuitively. Given $\EvalSet$ and $\OODSet$, we generate listener videos with a given attitude. We randomly select two sequences from each set and then down-sample to six frames to qualitatively visualize the generated results and the ground truth video. The results are shown in~\cref{fig:qualitative_results}, we can find that our model is generally able to capture listener patterns (\eg, \textit{eye}, \textit{mouth}, and \textit{head motion}, \etc.), which may differ from the ground-truth while still making sense.

From these two groups, we observe that the ``Random'' patterns behave very confusing and messy. The ``Simulation'' patterns depend heavily on whether we can randomize to a given attitude, and as long as this fails, the result is bad. The ``Simulation$^*$'' performs a little bit better, while its motion is also limited by the size and diversity of dataset, and intuitively, it cannot respond to the speaker in a dynamic manner. Our results are more visually plausible than others with head motions and expression changes.

We also provide the comparison results of listening head generation under different attitudes in~\cref{fig:three_attitude}. Obviously, the facial expression and head motions under different attitudes are expressive and distinguishable. 

\textbf{Ablation Studies}\quad We conduct an ablation study on the impact of different speaker signals for listening head modeling on $\EvalSet$. As~\cref{tab:diff_inputs} shown, the listening head driven by audio-only inputs prefers expression modeling but performs badly in head motion (angle, trans), while the model with visual-only inputs would capture the motion or sightline changes of the speaker during conversation and provide reasonable listening head action response. Removing these two modalities and simply mirror the speaker causes the worst listener. And with these two joint input signals, our model exhibits the best performance. That is, \emph{only when we look and hear, can we act as better responsive listeners}.

\setlength\intextsep{0em}
\begin{wraptable}[11]{r}{0.5\textwidth}
  \centering
  \small
  \caption{The averaged Feature Distance ($\times 100$) of listener generations across all attitudes on $\EvalSet$.}
  \vskip 1em
  \label{tab:diff_inputs}
  \begin{tabular}{@{}lccc@{}}
    \hline\noalign{\smallskip}
    Method      & angle & exp & trans \\
    \hline
    \noalign{\smallskip}
    Audio only  & \makecell[r]{8.69} & \makecell[r]{17.19} & \makecell[r]{8.49} \\
    Visual only & \makecell[r]{8.06} & \makecell[r]{18.85} & \makecell[r]{7.54} \\\hdashline[.4pt/1pt]
    Wrong Audio & \makecell[r]{8.15} & \makecell[r]{18.02} & \makecell[r]{7.60} \\\hdashline[.4pt/1pt]
    Mirroring & \makecell[r]{9.99} & \makecell[r]{26.48} & \makecell[r]{11.39} \\ \hdashline[.4pt/1pt]
    Ours    & \textbf{\makecell[r]{7.79}} & \textbf{\makecell[r]{15.04}} & \textbf{\makecell[r]{6.52}} \\
    \hline
    \end{tabular}
\end{wraptable}

Besides, we also report the performance of listener generation by giving wrong audio inputs and correct visual inputs in \cref{tab:diff_inputs}. It shows that Audio Only $<$ Wrong Audio $<$ Visual Only for expression FD, while Visual Only $\approx$ Wrong Audio $<$ Audio Only for head motion FD (lower is better). This also reveals that expression modeling depends more on audio inputs and head motion modeling is more corresponding to visual inputs.

\begin{figure}[t]
    \centering
    \includegraphics[width=\linewidth]{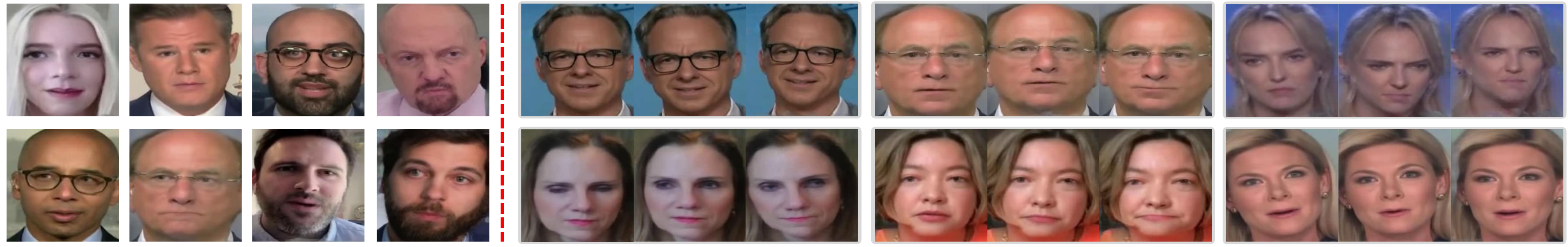}
    \caption{Diverse visual patterns can be observed from the generated listening-heads. Left: diverse expressions can be generated corresponding to different attitudes. Right: motion patterns including \textit{shaking head}, \textit{nod}, \textit{glare}, \textit{looking askance}, \textit{pressing lip} and \textit{focusing}}
    \label{fig:diversity}
\end{figure}

\textbf{Diversity of Generations}\quad \cref{fig:diversity} illustrates the diversity of visual patterns learned by our method. The single images on the left side show our model can generate different expressions while the six groups of images on the right demonstrate the head motions and eye contacts can be modeled by our method.

\textbf{Runtime Complexity}\quad With a given speaker's streaming video, it takes \SI{52.5}{\ms} to fit the 3DMM coefficients, and generate the next step listener's motion in \SI{0.0372}{\ms}, then render the motion to RGB images in \SI{29.3}{\ms}. The per-frame delay of this process on one Tesla-V100 GPU is \SI{81.84}{\ms} without any optimization strategy such as ONNX or TensorRT. The FPS of generated listening head videos can reach 12, and can be further improved to 19 with pipeline parallelism. The bottleneck of our proposed baseline method is the efficiency of 3D face reconstruction (fitting). Some video post-processing methods, such as video frame interpolation~\cite{huang2022rife,kong2022ifrnet,park2021asymmetric} can be used for real-time interaction.

\textbf{User study}\quad Since responsive listening head modeling is a user-oriented task, it is essential to conduct user studies for evaluation. For $\EvalSet$ and $\OODSet$, we had 10 volunteers doing two double-blind tests: 1) \textbf{Preference Test (PT).} Given the shuffled tuple of \textlangle ground-truth listening head, generated listening head\textrangle~, along with the ground-truth attitude, speaker's audio and speaker's video, the volunteers were asked to pick the best listening head. 2) \textbf{Attitude Matching Test (AttMatch).} Given the randomly shuffled listening heads generations conditioned by three different attitudes for each identity, the volunteers were asked to identify one attitude for each listening head. 

\begin{table}[t]
\setlength\abovecaptionskip{0\p@}
\setlength\belowcaptionskip{10\p@}
\centering
\caption{(a) The result (mean / variance) of preference test. ``Equal'': the generated results and the ground-truth are visually equivalent. ``PD Better'': the generated results are more in line with human perception. (b) Mean precision and precision variance of attitude matching test}
\begin{subtable}[h]{0.4\textwidth}
  \centering
  \setlength\abovecaptionskip{0\p@}
  \setlength\belowcaptionskip{0\p@}
  \caption{Preference Test}
  \label{tab:user_study_match}
  \begin{tabular}{@{}lr@{ / }rr@{ / }r@{}}
  \hline\noalign{\smallskip}
  PT & \multicolumn{2}{c}{$\EvalSet$} & \multicolumn{2}{c}{$\OODSet$} \\
  \noalign{\smallskip}
  \hline
  \noalign{\smallskip}
  Equal & 31.3 & 17.1 & 13.9 & 6.2 \\
  GT Better & 56.2 & 14.9 & 65.3 & 6.1 \\
  PD Better & 12.5 & 9.3 & 20.8 & 4.6 \\
  \hline
  \end{tabular}
\end{subtable}
~~~~~~~~~~
\begin{subtable}[h]{0.4\textwidth}
  \centering
  \setlength\abovecaptionskip{0\p@}
  \setlength\belowcaptionskip{0\p@}
  \caption{Attitude Matching Test}
  \label{tab:user_study_emo}
  \begin{tabular}{@{}lr@{ / }rr@{ / }r}
  \hline\noalign{\smallskip}
  AttMatch & \multicolumn{2}{c}{$\EvalSet$} & \multicolumn{2}{c}{$\OODSet$} \\
  \noalign{\smallskip}
  \hline
  \noalign{\smallskip}
  Positive & 66.7 & 15.7 & 69.0 & 7.1    \\
  Negative & 63.9 & 16.4 & 50.0 & 12.8   \\
  Neutral  & 69.4 & 9.2  & 53.6 & 3.9    \\
  \hline
  \end{tabular}
\end{subtable}
\end{table}

For the first user-study, each volunteer was asked to check 30 samples in a preference test. As shown in Tab.\ 6a, fpr $\EvalSet$ and $\OODSet$, volunteers voted that nearly 43.8\% and 44.7\% of the generated heads can get equal or even better rating than the ground-truth heads respectively, which verified that our model could generate responsive listeners consistent with subjective human perceptions, and even can be reached confused as real ones. 

Our second user-study results are shown in Tab.\ 6b. Each volunteer was given 90 generated listening clips under three different attitudes. For each attitude, we calculate the mean precision and precision variance across all volunteers for analysis. The results show that both in $\EvalSet$ and $\OODSet$, our model is capable of generating listening head yielding to the required attitude. The precision for negative and neutral attitude is slightly decreased in $\OODSet$, which might be caused by the unbalanced attitude distributions.
\section{Conclusion}
\label{sec:conclusion}
In this paper, we define the responsive listening head generation task. It aims to generate a responsive video clip for a listener with the understanding of the speaker's facial signals and voices. Further, the high-quality responsive listener dataset (ViCo) is contributed for addressing this problem. The responsive listener generation baseline can synthesis active listeners, which are more consistent with human perception. We expect that ViCo could benefit the face-to-face communication modeling in computer vision area and facilitate the applications in more scenarios, such as intelligence assistance, virtual human, \etc.

\noindent\textbf{Ethical Impact}\quad The ViCo dataset will be released only for research purposes under restricted licenses. The responsive listening patterns are identity-independent, which reduces the abuse of facial data. The only potential social harm is ``fake content''. However, different from talking head synthesis, responsive listening can hardly harm information fidelity.

\noindent\textbf{Acknowledgement} This work was supported by the National Key R\&D Program of China under Grant No. 2020AAA0108600.

\bibliographystyle{splncs04}
\bibliography{egbib}

\begin{thebibliography}{10}
\providecommand{\url}[1]{\texttt{#1}}
\providecommand{\urlprefix}{URL }
\providecommand{\doi}[1]{https://doi.org/#1}

\bibitem{afouras2018lrs3}
Afouras, T., Chung, J.S., Zisserman, A.: Lrs3-ted: a large-scale dataset for
  visual speech recognition. arXiv preprint arXiv:1809.00496  (2018)

\bibitem{bansal2018recycle}
Bansal, A., Ma, S., Ramanan, D., Sheikh, Y.: Recycle-gan: Unsupervised video
  retargeting. In: Proceedings of the European conference on computer vision
  (ECCV). pp. 119--135 (2018)

\bibitem{barker1971listening}
Barker, L.L.: Listening behavior.  (1971)

\bibitem{beltagy2020longformer}
Beltagy, I., Peters, M.E., Cohan, A.: Longformer: The long-document
  transformer. arXiv preprint arXiv:2004.05150  (2020)

\bibitem{berger2005interpersonal}
Berger, C.R.: Interpersonal communication: Theoretical perspectives, future
  prospects. Journal of communication  (2005)

\bibitem{blanz1999morphable}
Blanz, V., Vetter, T.: A morphable model for the synthesis of 3d faces. In:
  Proceedings of the 26th annual conference on Computer graphics and
  interactive techniques. pp. 187--194 (1999)

\bibitem{bohr2013no}
Bohr, P., Gargote, R., Vhorkate, R., Yawle, R., Bairagi, V.: A no reference
  image blur detection using cumulative probability blur detection (cpbd)
  metric. International Journal of Science and Modern Engineering
  \textbf{1}(5) (2013)

\bibitem{buschmeier2014alico}
Buschmeier, H., Malisz, Z., Skubisz, J., Wlodarczak, M., Wachsmuth, I., Kopp,
  S., Wagner, P.: Alico: A multimodal corpus for the study of active listening.
  In: LREC 2014, Ninth International Conference on Language Resources and
  Evaluation, 26-31 May, Reykjavik, Iceland. pp. 3638--3643 (2014)

\bibitem{cao2013facewarehouse}
Cao, C., Weng, Y., Zhou, S., Tong, Y., Zhou, K.: Facewarehouse: A 3d facial
  expression database for visual computing. IEEE Transactions on Visualization
  and Computer Graphics  \textbf{20}(3),  413--425 (2013)

\bibitem{cassel2000elements}
Cassel, N.N.W.W.: Elements of face-to-face conversation for embodied
  conversational agents, embodied conversational agents (2000)

\bibitem{chung2017you}
Chung, J.S., Jamaludin, A., Zisserman, A.: You said that? arXiv preprint
  arXiv:1705.02966  (2017)

\bibitem{chung2018voxceleb2}
Chung, J.S., Nagrani, A., Zisserman, A.: Voxceleb2: Deep speaker recognition.
  arXiv preprint arXiv:1806.05622  (2018)

\bibitem{chung2014empirical}
Chung, J., Gulcehre, C., Cho, K., Bengio, Y.: Empirical evaluation of gated
  recurrent neural networks on sequence modeling. arXiv preprint
  arXiv:1412.3555  (2014)

\bibitem{deng2019accurate}
Deng, Y., Yang, J., Xu, S., Chen, D., Jia, Y., Tong, X.: Accurate 3d face
  reconstruction with weakly-supervised learning: From single image to image
  set. In: IEEE Computer Vision and Pattern Recognition Workshops (2019)

\bibitem{fassaert2007active}
Fassaert, T., van Dulmen, S., Schellevis, F., Bensing, J.: Active listening in
  medical consultations: Development of the active listening observation scale
  (alos-global). Patient education and counseling  \textbf{68}(3),  258--264
  (2007)

\bibitem{gillies2008responsive}
Gillies, M., Pan, X., Slater, M., Shawe-Taylor, J.: Responsive listening
  behavior. Computer animation and virtual worlds  \textbf{19}(5),  579--589
  (2008)

\bibitem{ginosar2019learning}
Ginosar, S., Bar, A., Kohavi, G., Chan, C., Owens, A., Malik, J.: Learning
  individual styles of conversational gesture. In: Proceedings of the IEEE/CVF
  Conference on Computer Vision and Pattern Recognition. pp. 3497--3506 (2019)

\bibitem{hadar1985head}
Hadar, U., Steiner, T.J., Rose, F.C.: Head movement during listening turns in
  conversation. Journal of Nonverbal Behavior  \textbf{9}(4),  214--228 (1985)

\bibitem{heusel2017gans}
Heusel, M., Ramsauer, H., Unterthiner, T., Nessler, B., Hochreiter, S.: Gans
  trained by a two time-scale update rule converge to a local nash equilibrium.
  Advances in neural information processing systems  \textbf{30} (2017)

\bibitem{heylen2011generating}
Heylen, D., Bevacqua, E., Pelachaud, C., Poggi, I., Gratch, J., Schr{\"o}der,
  M.: Generating listening behaviour. In: Emotion-oriented systems, pp.
  321--347. Springer (2011)

\bibitem{heylen2007searching}
Heylen, D., Bevacqua, E., Tellier, M., Pelachaud, C.: Searching for
  prototypical facial feedback signals. In: International Workshop on
  Intelligent Virtual Agents. pp. 147--153. Springer (2007)

\bibitem{hochreiter1997long}
Hochreiter, S., Schmidhuber, J.: Long short-term memory. Neural computation
  \textbf{9}(8),  1735--1780 (1997)

\bibitem{homke2018eye}
H{\"o}mke, P., Holler, J., Levinson, S.C.: Eye blinks are perceived as
  communicative signals in human face-to-face interaction. PloS one
  \textbf{13}(12),  e0208030 (2018)

\bibitem{honeycutt2001mental}
Honeycutt, J.M., Ford, S.G.: Mental imagery and intrapersonal communication: A
  review of research on imagined interactions (iis) and current developments.
  Annals of the International Communication Association  \textbf{25}(1),
  315--345 (2001)

\bibitem{huang2022rife}
Huang, Z., Zhang, T., Heng, W., Shi, B., Zhou, S.: Real-time intermediate flow
  estimation for video frame interpolation. In: Proceedings of the European
  Conference on Computer Vision (ECCV) (2022)

\bibitem{jalongo1995promoting}
Jalongo, M.R.: Promoting active listening in the classroom. Childhood Education
   \textbf{72}(1),  13--18 (1995)

\bibitem{joo2019towards}
Joo, H., Simon, T., Cikara, M., Sheikh, Y.: Towards social artificial
  intelligence: Nonverbal social signal prediction in a triadic interaction.
  In: Proceedings of the IEEE/CVF Conference on Computer Vision and Pattern
  Recognition. pp. 10873--10883 (2019)

\bibitem{kendon1970movement}
Kendon, A.: Movement coordination in social interaction: Some examples
  described. Acta psychologica  \textbf{32},  101--125 (1970)

\bibitem{kendon2011organization}
Kendon, A., Harris, R.M., Key, M.R.: Organization of behavior in face-to-face
  interaction. Walter de Gruyter (2011)

\bibitem{kim2018deep}
Kim, H., Garrido, P., Tewari, A., Xu, W., Thies, J., Niessner, M., P{\'e}rez,
  P., Richardt, C., Zollh{\"o}fer, M., Theobalt, C.: Deep video portraits. ACM
  Transactions on Graphics (TOG)  \textbf{37}(4),  1--14 (2018)

\bibitem{kington2021identifying}
Kington, R.S., Arnesen, S., Chou, W.Y.S., Curry, S.J., Lazer, D., Villarruel,
  A.M.: Identifying credible sources of health information in social media:
  Principles and attributes. NAM perspectives  \textbf{2021} (2021)

\bibitem{kong2022ifrnet}
Kong, L., Jiang, B., Luo, D., Chu, W., Huang, X., Tai, Y., Wang, C., Yang, J.:
  Ifrnet: Intermediate feature refine network for efficient frame
  interpolation. In: Proceedings of the IEEE/CVF Conference on Computer Vision
  and Pattern Recognition. pp. 1969--1978 (2022)

\bibitem{li2021write}
Li, L., Wang, S., Zhang, Z., Ding, Y., Zheng, Y., Yu, X., Fan, C.:
  Write-a-speaker: Text-based emotional and rhythmic talking-head generation.
  In: Proceedings of the AAAI Conference on Artificial Intelligence. vol.~35,
  pp. 1911--1920 (2021)

\bibitem{loshchilov2017decoupled}
Loshchilov, I., Hutter, F.: Decoupled weight decay regularization. arXiv
  preprint arXiv:1711.05101  (2017)

\bibitem{luhmann1992communication}
Luhmann, N.: What is communication? Communication theory  \textbf{2}(3),
  251--259 (1992)

\bibitem{maatman2005natural}
Maatman, R., Gratch, J., Marsella, S.: Natural behavior of a listening agent.
  In: International workshop on intelligent virtual agents. pp. 25--36.
  Springer (2005)

\bibitem{mckeown2011semaine}
McKeown, G., Valstar, M., Cowie, R., Pantic, M., Schroder, M.: The semaine
  database: Annotated multimodal records of emotionally colored conversations
  between a person and a limited agent. IEEE transactions on affective
  computing  \textbf{3}(1),  5--17 (2011)

\bibitem{mcnaughton2008learning}
McNaughton, D., Hamlin, D., McCarthy, J., Head-Reeves, D., Schreiner, M.:
  Learning to listen: Teaching an active listening strategy to preservice
  education professionals. Topics in Early Childhood Special Education
  \textbf{27}(4),  223--231 (2008)

\bibitem{melis2019mogrifier}
Melis, G., Ko{\v{c}}isk{\`y}, T., Blunsom, P.: Mogrifier lstm. arXiv preprint
  arXiv:1909.01792  (2019)

\bibitem{mineyama2007supervisors}
Mineyama, S., Tsutsumi, A., Takao, S., Nishiuchi, K., Kawakami, N.:
  Supervisors' attitudes and skills for active listening with regard to working
  conditions and psychological stress reactions among subordinate workers.
  Journal of occupational health  \textbf{49}(2),  81--87 (2007)

\bibitem{oertel2021towards}
Oertel, C., Jonell, P., Kontogiorgos, D., Mora, K.F., Odobez, J.M., Gustafson,
  J.: Towards an engagement-aware attentive artificial listener for multi-party
  interactions. Frontiers in Robotics and AI p.~189 (2021)

\bibitem{park2021asymmetric}
Park, J., Lee, C., Kim, C.S.: Asymmetric bilateral motion estimation for video
  frame interpolation. In: Proceedings of the IEEE/CVF International Conference
  on Computer Vision. pp. 14539--14548 (2021)

\bibitem{parker2000improving}
Parker, J., Coiera, E.: Improving clinical communication: a view from
  psychology. Journal of the American Medical Informatics Association
  \textbf{7}(5),  453--461 (2000)

\bibitem{bfm09}
Paysan, P., Knothe, R., Amberg, B., Romdhani, S., Vetter, T.: A 3d face model
  for pose and illumination invariant face recognition. In: 2009 sixth IEEE
  international conference on advanced video and signal based surveillance. pp.
  296--301. Ieee (2009)

\bibitem{petridis2013mahnob}
Petridis, S., Martinez, B., Pantic, M.: The mahnob laughter database. Image and
  Vision Computing  \textbf{31}(2),  186--202 (2013)

\bibitem{prajwal2020lip}
Prajwal, K., Mukhopadhyay, R., Namboodiri, V.P., Jawahar, C.: A lip sync expert
  is all you need for speech to lip generation in the wild. In: Proceedings of
  the 28th ACM International Conference on Multimedia. pp. 484--492 (2020)

\bibitem{ramamoorthi2001efficient}
Ramamoorthi, R., Hanrahan, P.: An efficient representation for irradiance
  environment maps. In: Proceedings of the 28th annual conference on Computer
  graphics and interactive techniques. pp. 497--500 (2001)

\bibitem{ramamoorthi2001signal}
Ramamoorthi, R., Hanrahan, P.: A signal-processing framework for inverse
  rendering. In: Proceedings of the 28th annual conference on Computer graphics
  and interactive techniques. pp. 117--128 (2001)

\bibitem{ren2021pirenderer}
Ren, Y., Li, G., Chen, Y., Li, T.H., Liu, S.: Pirenderer: Controllable portrait
  image generation via semantic neural rendering. In: Proceedings of the
  IEEE/CVF International Conference on Computer Vision. pp. 13759--13768 (2021)

\bibitem{richard2021meshtalk}
Richard, A., Zollh{\"o}fer, M., Wen, Y., De~la Torre, F., Sheikh, Y.: Meshtalk:
  3d face animation from speech using cross-modality disentanglement. In:
  Proceedings of the IEEE/CVF International Conference on Computer Vision. pp.
  1173--1182 (2021)

\bibitem{robertson2005active}
Robertson, K.: Active listening: more than just paying attention. Australian
  family physician  \textbf{34}(12) (2005)

\bibitem{rogers1957active}
Rogers, C.R., Farson, R.E.: Active listening (1957)

\bibitem{rost2013active}
Rost, M., Wilson, J.: Active listening. Routledge (2013)

\bibitem{stacks2014integrated}
Stacks, D.W., Salwen, M.B.: An integrated approach to communication theory and
  research. Routledge (2014)

\bibitem{sutskever2014sequence}
Sutskever, I., Vinyals, O., Le, Q.V.: Sequence to sequence learning with neural
  networks. In: Advances in neural information processing systems. pp.
  3104--3112 (2014)

\bibitem{tomasello2010origins}
Tomasello, M.: Origins of human communication. MIT press (2010)

\bibitem{vaswani2017attention}
Vaswani, A., Shazeer, N., Parmar, N., Uszkoreit, J., Jones, L., Gomez, A.N.,
  Kaiser, {\L}., Polosukhin, I.: Attention is all you need. Advances in neural
  information processing systems  \textbf{30} (2017)

\bibitem{wang2020mead}
Wang, K., Wu, Q., Song, L., Yang, Z., Wu, W., Qian, C., He, R., Qiao, Y., Loy,
  C.C.: Mead: A large-scale audio-visual dataset for emotional talking-face
  generation. In: European Conference on Computer Vision. pp. 700--717.
  Springer (2020)

\bibitem{wang2004image}
Wang, Z., Bovik, A.C., Sheikh, H.R., Simoncelli, E.P.: Image quality
  assessment: from error visibility to structural similarity. IEEE transactions
  on image processing  \textbf{13}(4),  600--612 (2004)

\bibitem{wu2018reenactgan}
Wu, W., Zhang, Y., Li, C., Qian, C., Loy, C.C.: Reenactgan: Learning to reenact
  faces via boundary transfer. In: Proceedings of the European conference on
  computer vision (ECCV). pp. 603--619 (2018)

\bibitem{zhang20213d}
Zhang, C., Ni, S., Fan, Z., Li, H., Zeng, M., Budagavi, M., Guo, X.: 3d talking
  face with personalized pose dynamics. IEEE Transactions on Visualization and
  Computer Graphics  (2021)

\bibitem{zhang2021facial}
Zhang, C., Zhao, Y., Huang, Y., Zeng, M., Ni, S., Budagavi, M., Guo, X.:
  Facial: Synthesizing dynamic talking face with implicit attribute learning.
  In: Proceedings of the IEEE/CVF International Conference on Computer Vision.
  pp. 3867--3876 (2021)

\bibitem{zhang2016joint}
Zhang, K., Zhang, Z., Li, Z., Qiao, Y.: Joint face detection and alignment
  using multitask cascaded convolutional networks. IEEE Signal Processing
  Letters  \textbf{23}(10),  1499--1503 (2016)

\bibitem{zhu2021deep}
Zhu, H., Luo, M.D., Wang, R., Zheng, A.H., He, R.: Deep audio-visual learning:
  A survey. International Journal of Automation and Computing pp. 1--26 (2021)

\end{thebibliography}

\clearpage
\newpage
\appendix
\renewcommand{\thetable}{A\arabic{table}}
\renewcommand{\thefigure}{A\arabic{figure}}
\setcounter{figure}{0}    
\setcounter{table}{0}

\section{Dataset Details} \label{sec:supp_dataset_detail}
\subsection{Clip Length / Duration Distribution}
\begin{figure}[h]
    \centering
    \includegraphics[width=0.9\linewidth,page=1]{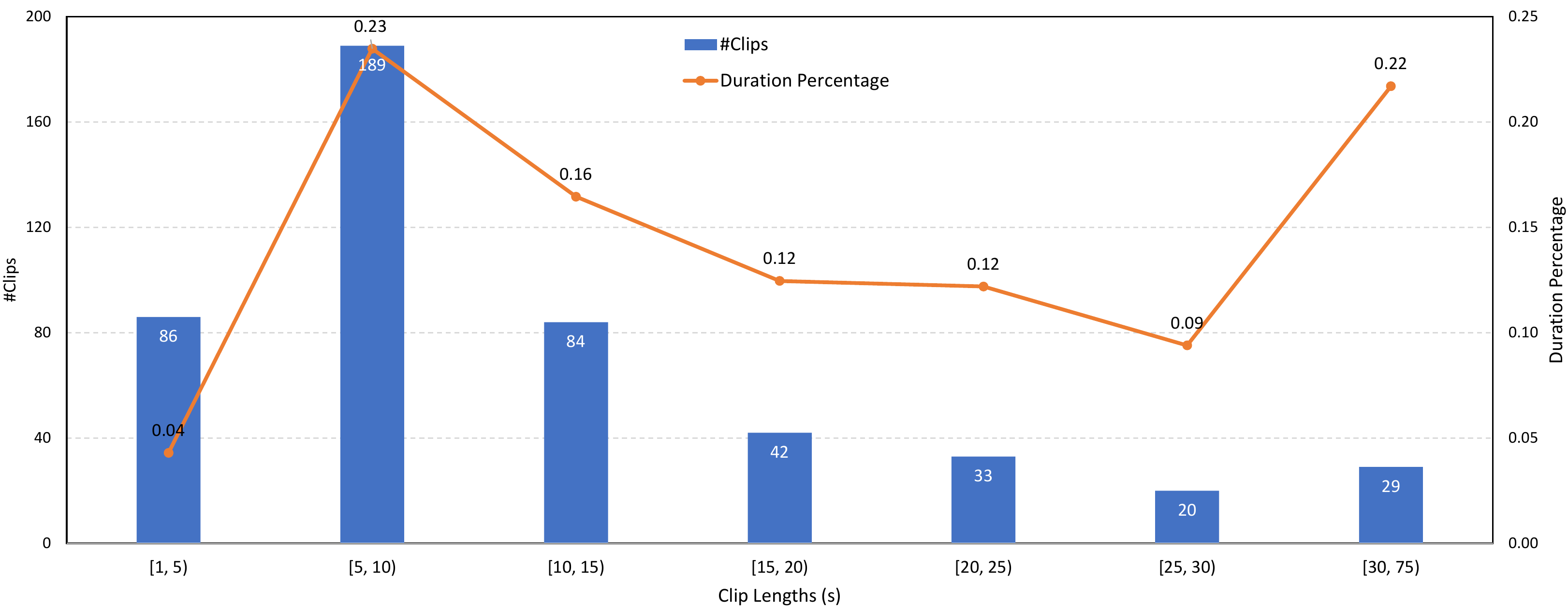}
    \caption{The clip length distribution and the corresponding duration percentage in ViCo dataset.}
    \label{fig:supp_clip_length_dist}
\end{figure}

We counted the distribution of clip lengths and the corresponding duration percentage in ViCo, and the results are shown in~\cref{fig:supp_clip_length_dist}. 

\subsection{Identities}
\begin{figure}[h]
    \centering
    \includegraphics[width=0.95\linewidth]{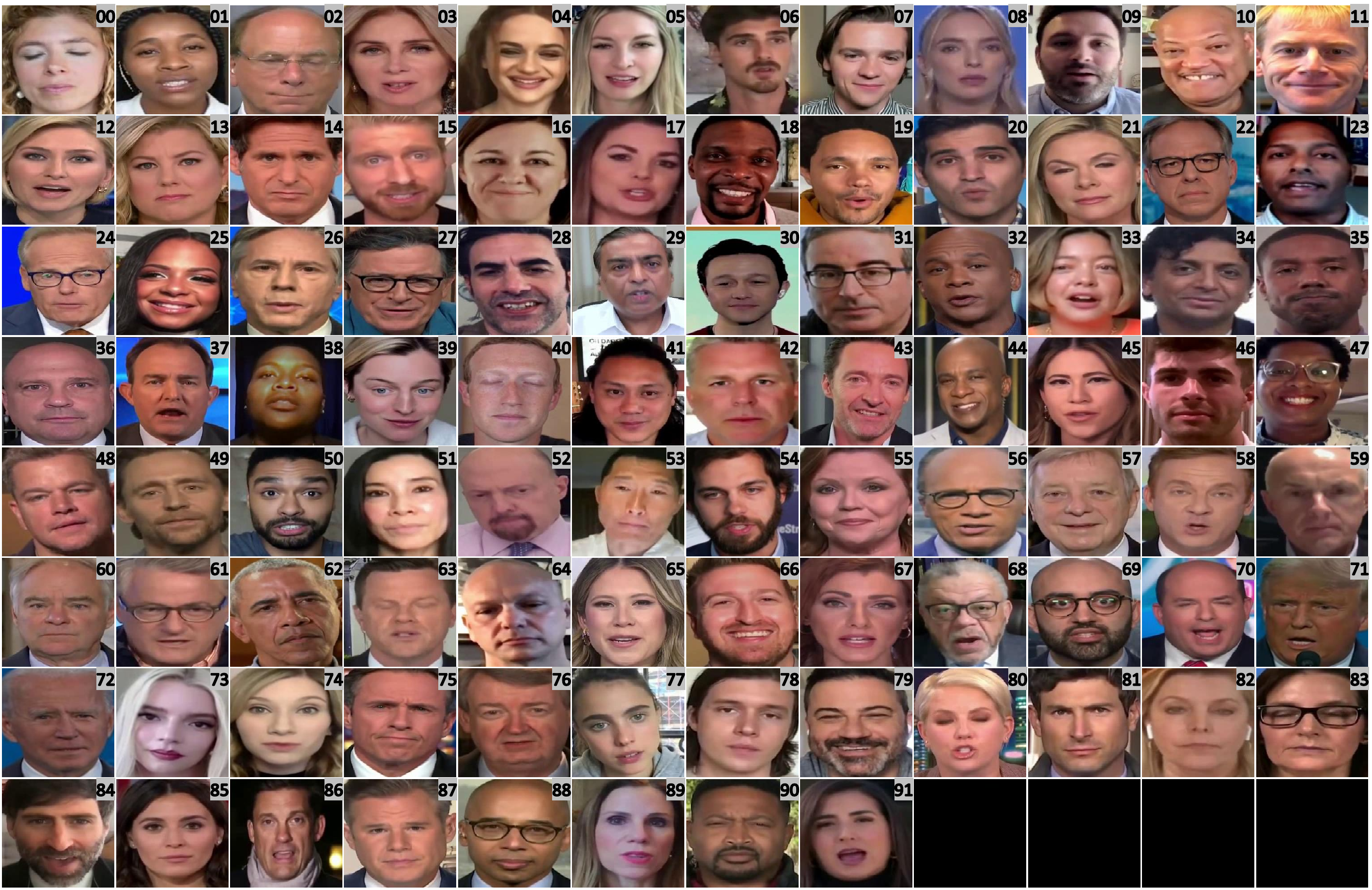}
    \caption{Ninty-two identities with different genders and races in our dataset.}
    \label{fig:supp_identity}
\end{figure}

The ViCo dataset contains 92 identities. In the~\cref{fig:supp_identity}, we picked a random image for each identity. It can be found that there are different genders and races; thus, the potential ethics problems can be reduced, \textit{i.e.}, the dataset represent the diversity of the community.

\begin{figure}[h]
    \centering
    \includegraphics[width=\linewidth,page=1]{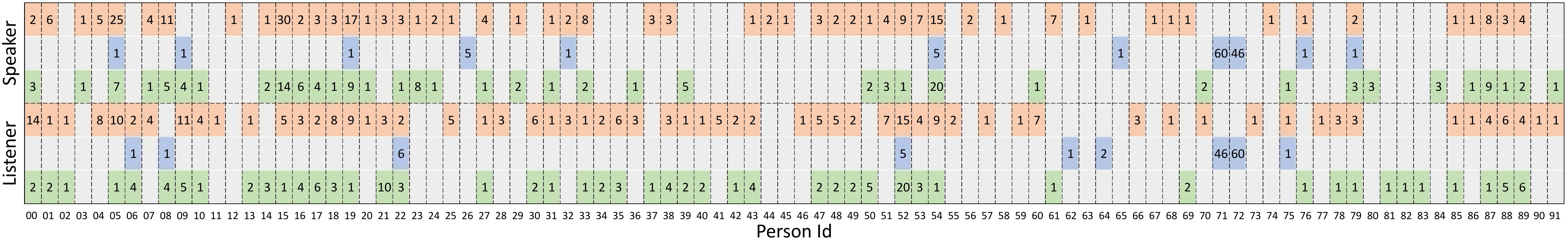}
    \caption{The identity distribution in our dataset. If an identity in a certain role (speaker/listener) shows a certain attitude (\textcolor{apositive}{positive}, \textcolor{anegative}{negative} and \textcolor{aneutral}{neutral}), then the corresponding cell is colored, and the number of clips is also attached.}
    \label{fig:supp_identity_dist}
\end{figure}

Some identities in our dataset may act as different roles (speaker/listener) and present different listener patterns under different attitudes. So, we statistic the identity distribution in~\cref{fig:supp_identity_dist}. This demonstrates the diversity of our dataset.

\section{IRB Approval} \label{sec:supp_irb}
The YouTube community has a strict censorship mechanism to avoid violent or dangerous content~\cite{kington2021identifying}. And as shown in the copyright\footnote{\url{https://www.youtube.com/howyoutubeworks/policies/copyright/\#fair-use}}, research purpose is considered fair use, which is allowed the reuse of copyright-protected material without getting permission from the copyright owner. This guarantees our dataset is congruence with the ethics guidelines.

\section{Pipeline Details} \label{sec:supp_experiment}
\subsection{3DMM Coefficients}
We extract the 3DMM coefficients following~\cite{deng2019accurate,ren2021pirenderer}, with the commonly used toolkit\footnote{\url{https://github.com/microsoft/Deep3DFaceReconstruction}} and the guides of PIRender\footnote{\url{https://github.com/RenYurui/PIRender}}, we can obtain a parametric representation of the face: $\{\alpha, \beta, \delta, p, \gamma\}$ which denote the identity, expression, texture, pose and lighting, respectively. Here, $\alpha\in\bR^{80}, \beta\in\bR^{64}, \delta\in\bR^{80}$, and $\gamma\in\bR^{27}$ for RGB channels in three-bands Spherical Harmonics~\cite{ramamoorthi2001efficient,ramamoorthi2001signal} representations, $p\in\bR^{6}$ to represent rotations with SO(3)$\in\bR^3$ and translations in $\bR^3$.

Therefore, the relative fixed, identity-dependent features $\cI=(\alpha, \delta, \gamma)$ is in $\bR^{187}$, and the relative dynamic, identity-independent features $m=(\beta, p)$ is in $\bR^{70}$. Additionally, to better model the head movements and make it compatible with PIRender~\cite{ren2021pirenderer}, we use a new ``crop'' parameter of $\bR^{3}$ in practice. This guides where we will place and size the parametric 3D face in the original image.

\begin{figure}[t]
    \centering
    \includegraphics[height=0.6\linewidth,page=1]{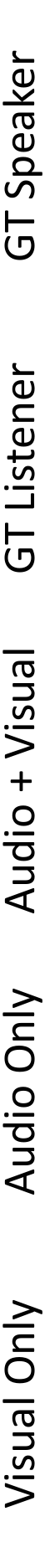}
    \begin{subfigure}[t]{0.36\linewidth}
        \centering
        \includegraphics[width=\textwidth,page=2]{figures/diff_inputs_v1_abc.pdf}
        \caption{}
        \label{fig:diff_inputs_test}
    \end{subfigure}
    \begin{subfigure}[t]{0.36\linewidth}
        \centering
        \includegraphics[width=\textwidth,page=3]{figures/diff_inputs_v1_abc.pdf}
        \caption{}
        \label{fig:diff_inputs_ood}
    \end{subfigure}
    \caption{Qualitative comparisons of listening head generation with different speaker signals (shown in rows). The attitude is positive for all listening heads}
    \label{fig:diff_inputs}
\end{figure}\textbf{}

\subsection{Model Complexity}
Our proposed model is lightweight (248K params) and efficient (941K flops for 30 frames input) with the 3DMM coefficients.

\section{Additional Experimental Results} \label{sec:supp_ablation}
\textbf{Qualitative Results with Different Modality Inputs}\quad We conduct an ablation study on the impact of different speaker signals for listening head modeling. The qualitative results are shown in~\cref{fig:diff_inputs}. We sample three frames from $\EvalSet$ (\cref{fig:diff_inputs_test}) and $\OODSet$ (\cref{fig:diff_inputs_ood}) to make a simplified but clear visualization. From~\cref{fig:diff_inputs_test}, 
show that models with only a single input have less head motion and expression variations, and furthermore, the audio only model is able to express some expressions (columns 1, 2) as full input model, while the video only model performs even worse in expression modeling, which may indicate that the speaker's audio signals contributes more on listener's expression. From~\cref{fig:diff_inputs_ood}, we can find that the listener is more likely to lose focus in the absence of video input (columns 2, 3), which may indicate that the speaker's visual signals can guide the listener where to focus.

\begin{table}[]
  \centering
  \small
  \caption{The Feature Distance ($\times 100$) of generations with different modality inputs and architectures \textbf{across all attitudes} (Averaged) on $\EvalSet$.}
  \label{tab:rebuttal_ablation}
  \begin{tabular}{@{}lccc@{}}
    \hline
    Method      & angle & exp & trans \\
    \hline
    Audio only  & \makecell[r]{8.69} & \makecell[r]{17.19} & \makecell[r]{8.49} \\
    Visual only & \makecell[r]{8.06} & \makecell[r]{18.85} & \makecell[r]{7.54} \\\hdashline[.4pt/1pt]
    Non-sequential Model & \makecell[r]{8.40} & \makecell[r]{18.50} & \makecell[r]{7.06} \\
    \hdashline[.4pt/1pt]
    Ours    & \textbf{\makecell[r]{7.79}} & \textbf{\makecell[r]{15.04}} & \textbf{\makecell[r]{6.52}} \\
    \hline
    \end{tabular}
\end{table}

\begin{figure}[h]
    \centering
    \includegraphics[width=0.8\linewidth]{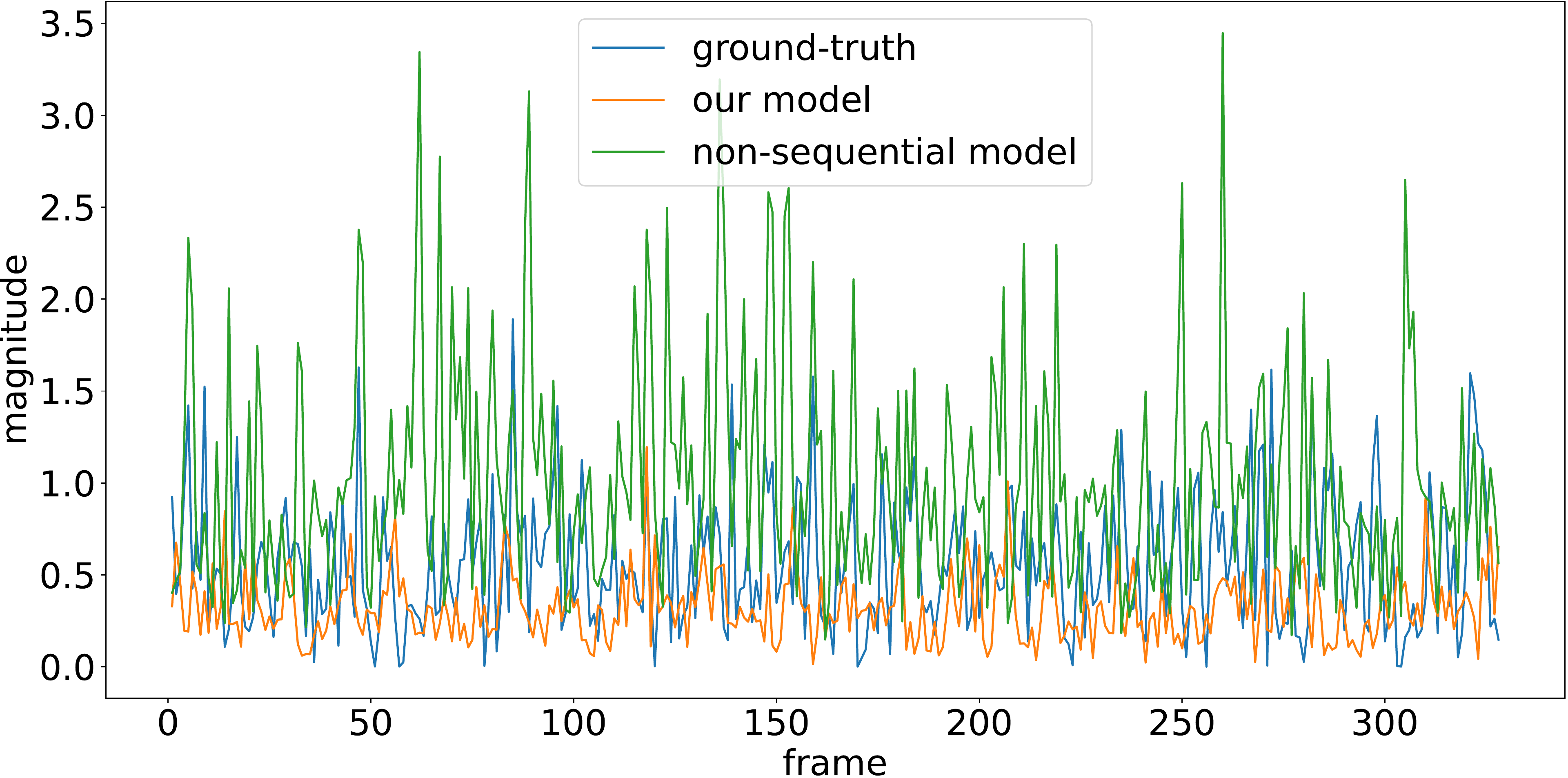}
    \caption{The magnitude of variation for feature {\tt angle} and {\tt trans} between frame $t$ and frame $t+1$ with $L_1$ distance}
    \label{fig:parallel_visualize}
\end{figure}

\noindent\textbf{Comparisons with Non-sequential Model}\quad Furthermore, we also experimented generate the listener frames in a ``purely parallel (non-sequential) manner'' rather than our ``auto-regressive decoder manner'', by removing the temporal connections and converting the LSTM cells to fully-connected layers with similar \#params. The results are shown in~\cref{tab:rebuttal_ablation} which worse than our model, since the ``non-sequential model'' results in noise and unnatural head motion caused by the lack of temporal constraint, as shown in ~\cref{fig:parallel_visualize}.

\section{Applications and Limitations} \label{sec:supp_applications}
\textbf{Applications}\quad Active listening faces ineffective communication and plays an important role in human-to-human interaction. People are encouraged to learn to actively listen to others in many scenarios, such as doctor - patient, teacher - student, salesperson - customer, \etc. Our proposed responsive listening head generation task fills a gap in modeling face-to-face communication in the computer vision area. It can be applied to many scenarios, such as human-computer interaction, intelligence assistance, virtual human, \etc. It would contribute to the mutual interaction between the virtual human and the real human. It can also be adopted to virtual audience modeling or providing guidance about how to act as an active listener. 

\noindent\textbf{Limitations}\quad One potential limitation of our constructed dataset is that the attitude is assumed to be consistent across the clips, since we cut and annotate short clips from the candidate video. However, as shown in the qualitative results, the generated head sequences can vary from one common head respecting different attitudes. We may deduce that our model is feasible to the attitude-transferable conversations.

\end{document}